%% file: semeval2020.tex
\documentclass[11pt]{article}
\usepackage{geometry}
\usepackage{coling2020}
\usepackage{times}
\usepackage{url}
\usepackage{latexsym}
\usepackage{microtype}
\usepackage{pgfplots}
\usepackage{multirow}
\usepackage{float}
\usepackage{textcomp}

\usepackage{amsmath,amssymb}

\colingfinalcopy % Uncomment this line for all SemEval submissions

\title{JokeMeter at SemEval-2020 Task 7: Convolutional humor }

\author{Martin Docekal, Martin Fajcik, Josef Jon, Pavel Smrz\\
  Brno University of Technology, Faculty of Information Technology\\
  612\,66 Brno, Czech Republic \\
  {\tt \{idocekal,ifajcik,ijon,smrz\}@fit.vutbr.cz} }

\date{}

\begin{document}
\maketitle
\begin{abstract}
This paper describes our system that was designed for Humor evaluation within the SemEval-2020 Task~7. The system is based on convolutional neural network architecture. We investigate the system on the official dataset, and we provide more insight to model itself to see how the learned inner features look.
\end{abstract}

\section{Introduction}
\label{intro}

\begin{figure}[h]
    \centering
    \includegraphics[width=0.7\linewidth]{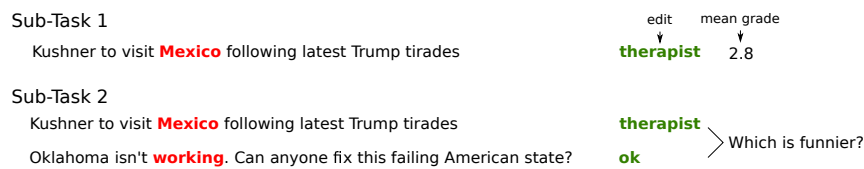}

    \caption{\label{fig:subtasksExamples} Examples for both tasks. The red color in the original title marks part that is substituted with a green word.}
    
\end{figure}

This paper deals with estimating the humor of edited  English news headlines~\cite{hossain-etal-2020-funlines,SemEval2020Task7}. The illustration of tasks is in Figure \ref{fig:subtasksExamples}. The original text sequence is given, which represents a title, with the annotated part that is edited along with the edit itself. Our responsibility is to determine how funny this change is in the range from 0 to 3 (inclusive). This is called Sub-Task 1. We also participate in the Sub-Task 2, in that we should decide which from both given edits is the funnier one.  For the second task, we used the approach of reusing the model from the first task as it is described in section \ref{jokeMeter}. So we are focusing the description on Sub-Task 1.

Official results were achieved with a Convolutional Neural Networks (CNNs) \cite{lecun1999object,fukushima1982neocognitron}, but we also tested numerous other approaches such as SVM~\cite{SVM} and pre-trained transformer model~\cite{transformers}.

The humor is a very subjective phenomenon, as can be seen from the inter-agreement on label annotation in Sub-Task 1 dataset\footnote{The inter-annotator agreement measured with Krippendorff's interval metric is just 0.2~ \cite{hossain-etal-2019-president}.}. The given data labels do not allow us to learn a sense of humor of a human annotator because the dataset does not specify from whom the grade comes. So, for example, if we have one annotator that likes dark humor and all the others not, we will be considering such a replacement as not humorous no meter if it is excellent dark humor or not. In other words, we may say that we are searching for some most common kind of humor.

The dominant theory of humor is the Incongruity Theory \cite{sep-humor}. It says that we are finding humor in perceiving something unexpected (incongruous) that violates expectations that were set up by the joke. 
There are samples, in the provided dataset, that uses the incongruity to create humor. Moreover, according to  Hossain et al.  \shortcite{SemEval2020Task7}, we can see a positive influence of incongruity on systems results for the dataset.

%
% The following footnote without marker is needed for the camera-ready
% version of the paper.
% Comment out the instructions (first text) and uncomment the 8 lines
% under "final paper" for your variant of English.
% 
\blfootnote{
    %
    % for review submission
    %
    %\hspace{-0.65cm}  % space normally used by the marker
    % Place licence statement here for the camera-ready version. See
    %Section~\ref{licence} of the instructions for preparing a
    %manuscript.
    %
    % % final paper: en-uk version 
    %
    % \hspace{-0.65cm}  % space normally used by the marker
    % This work is licensed under a Creative Commons 
    % Attribution 4.0 International Licence.
    % Licence details:
    % \url{http://creativecommons.org/licenses/by/4.0/}.
    % 
    % % final paper: en-us version 
    %
    \hspace{-0.65cm}  % space normally used by the marker
This work is licensed under a Creative Commons Attribution 4.0 International Licence. Licence details: \url{http://creativecommons.org/licenses/by/4.0/}.
}

\section{Related Work}
\label{relatedWork}

Computational humor is usually divided into two groups: recognition and generation. Humor generation focuses on creating the humor itself, so it can be a system that is able to tell a joke.

Recognition can be a binary (funny or not) classification task, but in our work, we also want to know how funny a sequence is by rating it with a grade in a given interval. Hossain \shortcite{hossain-etal-2019-president}, which introduced the dataset used in this work, tried to create models that can classify if a given edited title is funny or not, so they were training just binary classifiers, in contrast with our regression approach.

Many other works use this binary approach \cite{cattle-ma-2018-recognizing,chen-soo-2018-humor,bertero-fung-2016-long,weller-seppi-2019-humor,purandare-litman-2006-humor}. They use various models, such as LSTM \cite{LSTM}, CNN, and BERT \cite{bert}, to deal with this task.

Our official results were achieved with a CNN model that is inspired by architecture presented in \cite{zhang2015sensitivity}. Their architecture is compact to its size (one layer CNN), so it provides the advantage of using less computational resources, in contrast with big models like BERT. Even with the small size, they were able to achieve promising results for the sentence classification task. Also, such a small model allows us to gain better insight into what is going on underneath.

\section{Data}
\label{data}

In this section, we would like to point out some interesting facts about the used data. We are focusing on the data for Sub-Task 1 because we used the model trained on the Sub-Task 1 for Sub-Task 2 (more in section \ref{jokeMeter}).

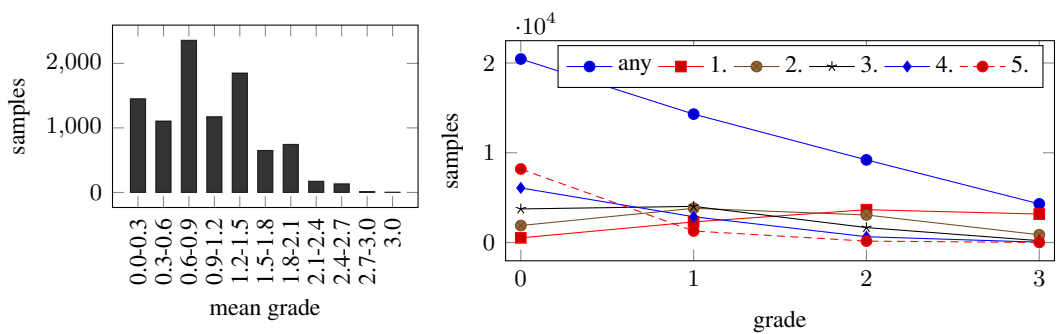
\begin{figure}[h]
    \centering
    \begin{minipage}{.35\textwidth}
      \input{images/histogram_mean_imbalanced.tex}
    \end{minipage}
    \begin{minipage}{.55\textwidth}
      \input{images/imbalancement_per_position.tex}
    \end{minipage}
    \caption{\label{fig:histTrainImbalance}In the histogram on the left side, we can see the number of samples in the train set that have a mean grade in a given bin. The bins are left-inclusive. The graph on the right side shows the number of grades of a given type in the whole train set if we always take just grade on given position per sample.}
    
\end{figure}

For each example, the dataset \cite{hossain-etal-2019-president} provides annotation in the form of grades (0,1,2 and 3) of humor assessment in sorted descending order and the mean of these grades. In most cases, there are five grades per dataset sample (sometimes more). In our work, we always use the first five grades. As can be seen from graphs in Figure \ref{fig:histTrainImbalance}, the dataset is imbalanced, and the high grades are rare. Also, we investigated how the dataset is imbalanced when considering just single n-th grade and omit the others. Though we can say, from the graph on the right side of the very same figure, that there is still imbalance, we can see that mainly the 2. and 3. positions seem to have smaller fluctuance among the number of samples per grade than the other positions. 

\begin{table}
\begin{minipage}{.40\linewidth}

    \centering
    \begin{tabular}{|r|c|c|c|c|c|}
    \hline
    \textbf{position} & 1     & 2     & 3     & 4    & 5     \\ \hline
    \textbf{RMSE}     & 1.179 & 0.583 & 0.403 & 0.63 & 0.903 \\ \hline
    \end{tabular}
    \caption{\label{tab:tableOfPositionsRes}Results on the Sub-Task 1 train set of the oracle classifier, which always correctly predict the grade on the n-th position.}
    
\end{minipage}\hspace{2.7cm}
\begin{minipage}{.40\linewidth}
    \centering
    \begin{tabular}{|r|c|c|c|c|}
    \hline
    \textbf{grade} & 0     & 1     & 2     & 3     \\ \hline
    \textbf{RMSE}  & 1.103 & 0.587 & 1.214 & 2.145 \\ \hline
    \end{tabular}
    \caption{\label{tab:tableOfSameGradeRes}This table shows results for a classifier that would always predict the same grade for the Sub-Task 1 train set. }
\end{minipage}
\end{table}

We also did further analysis to determine prediction quality for the case when we would have an oracle classifier always predicting the grade on the \textit{n-th} position. The results are in Table \ref{tab:tableOfPositionsRes}. It can be seen that the third position is superior.

Another thing that we decided to investigate is how the RMSE score would look like if we always predict the same grade. The results are in Table \ref{tab:tableOfSameGradeRes}. 

\section{System Description}
\label{jokeMeter}

\begin{figure}[h]
    \centering
    \includegraphics[width=1.0\linewidth]{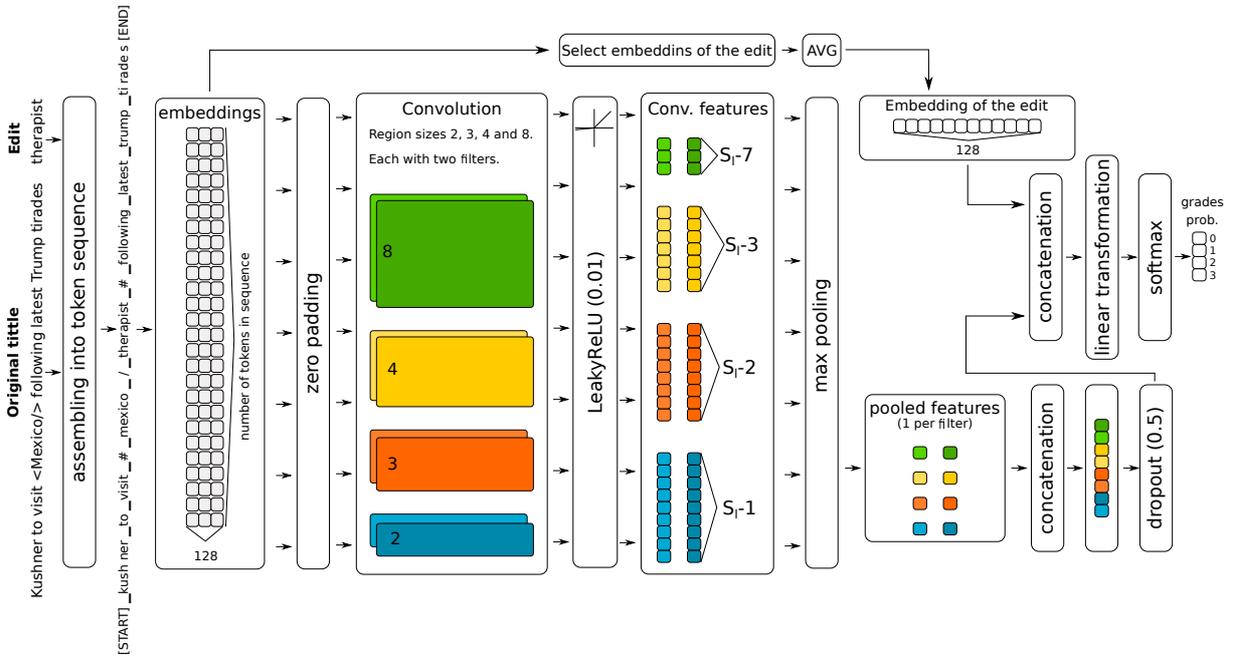}

    \caption{\label{fig:jokeMeterModel} Illustration of our submission model. The $S_l$ is a length of sequence that is processed by the convolution filters.}
    
\end{figure}

The main inspiration for our model architecture (JokeMeter) comes from Zhang and Wallace's \shortcite{zhang2015sensitivity} work. The used model has CNN architecture illustrated in Figure \ref{fig:jokeMeterModel}. Firstly the input sequence is assembled. The edit is inserted into the original title after the part that is being edited. Additionally, we add a slash separator, and the whole original/edit location is delimited with two hashtags. In this way, we were able to add input for the model that has complete information about the original and the edited title. We also include tokens to mark the start and the end of a title. The reason behind these tokens is that we want to encode information into an n-gram, whether it is from the beginning/end of a title or not, and possibly make it easier for the model to learn the setup and punchline humor.

We tokenized the input by ALBERT  \cite{albert} pre-trained SentencePiece \cite{sentencepiece} tokenizer. Each token is assigned a 128-dimensional embedding from 30 000 size vocabulary. Right before the convolution, is added zero-padding of size one on both sides of the sequence. Each sequence is 512 tokens long; shorter sequences are padded with unique padding tokens.

We used four different convolution filter region sizes 2, 3, 4, and 8. For each size, we had two filters. These filters are followed with LeakyReLu \cite{leakyRelu} activation (negative slope is 0.01). We also experimented with a model (JokeMeterBoosted) that uses 2048 filters for each size, 2048-dimensional embeddings, and does not use the embedding of the edit.

In the final part of our model, we do max pooling in order to get one feature per filter (8 in total). We concatenate the features to compose a vector, and then that vector is again concatenated with the edit embedding. The edit embedding is an average of embeddings of all tokens the edit is composed of.

On the very end, we perform linear transformation followed by the softmax to get probabilities for each grade. With that configuration, we would not be doing a regression task, so at the test time we must do one final calculation that will transform these probabilities into a grade from the continuous interval [0,3]:

\begin{equation}
    \mathbb{E}[G|S=s]=\sum_{i=0}^{i=3} p_i i \, .
\end{equation}{}

Where the $G$ is a grade, $s$ is a input sequence, and $p_i$ is the estimated probability for grade $i$.  In the case of Sub-Task 2, we run the model for both titles separately, and in the end, we made the decision by comparing their estimated grades.

\subsection{Convolutional features analysis}

\begin{figure}[h]
    \centering
    \includegraphics[width=0.8\linewidth]{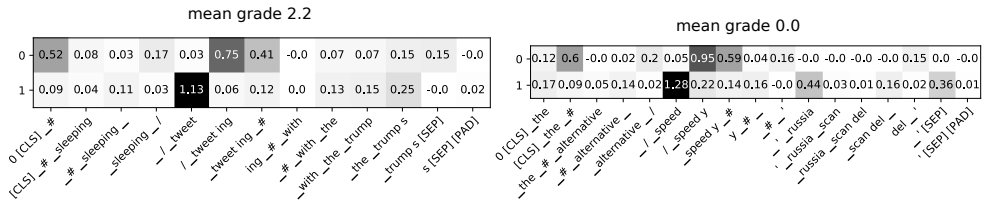}

    \caption{\label{fig:realExamplesOfConvFeatures} Real word examples of convolved features for trigrams.}
    
\end{figure}

We used two filters for each region size because we expected that the model would be able to train one feature that will signalize funny and one feature that will signalize not funny. Nevertheless, our analysis instead shows, it learns features that just signalizes how much not funny the given n-gram is, as can be seen in Figure \ref{fig:realExamplesOfConvFeatures}. To gain further insight into this property, we calculated, for the Sub-Task 1 train set, Spearman's rank correlation coefficient between mean grade and each feature (after max pooling). The results in Table \ref{tab:convFeatureAnal} shows a negative correlation that corresponds to our hypothesis. An interesting finding is that quite a lot of features have zero variance, which means that constant was learned. That leads us to think that these features reflect the fact that we are able to achieve relatively good RMSE with predicting constant (e.g., one as shown in Table \ref{tab:tableOfSameGradeRes}) due to the imbalance in the dataset.

\begin{table}[]
\centering
\begin{tabular}{|r|c|c|c|c|c|c|c|c|}
\hline
\textbf{feature}     & \multicolumn{4}{c|}{\textbf{0}}                   & \multicolumn{4}{c|}{\textbf{1}}                   \\ \hline
\textbf{region size} & \textbf{2} & \textbf{3} & \textbf{4} & \textbf{8} & \textbf{2} & \textbf{3} & \textbf{4} & \textbf{8} \\ \hline
\boldmath$\sigma$           & 0          & 0.22       & 0          & 0.21       & 0          & 0.13       & 0.23       & 0          \\ \hline
\textbf{r\textsubscript{s}}          & -          & -0.43      & -          & -0.52      & -          & -0.6       & -0.44      & -          \\ \hline
\end{tabular}
\caption{\label{tab:convFeatureAnal} Table of Spearman's rank correlation ($n=38608$, $p < 0.01$) coefficients between mean grade and each feature. Also, standard deviations of features values.}
\end{table}

\section{Experiments}
\label{experiments}

In this section, we describe the experiments we performed, not just with the model that is described in section \ref{jokeMeter}. Apart from models based on neural networks, we compiled several baselines: Decision Tree Classifier (DTC) \cite{breiman1984classification}, SVM, k-NN, and Naive Bayes Classifier (NBC). Also, we experimented with a model that uses transformer architecture (ALBERT-base-v2). 

The neural models were implemented with PyTorch \cite{pytorch}. For the ALBERT and the tokenizer, we used Hugging Face \cite{HuggingFacesTS} implementations, and for non-neural models, the implementations from the scikit-learn \cite{scikit-learn} were used.

We did two kinds of experiments for all models.  The first kind uses all five grades for each sample (\textit{all-grades} training) during the training. Every sample is copied five times, and a grade is assigned to each of them, so we have five samples that have the same content, but each can have a different grade. On the other hand, the second kind of experiment uses only the 3rd grade (\textit{3-grade} training), which has for oracle classifier the best score (see Table \ref{tab:tableOfPositionsRes}).

\subsection{Non-neural models}
\label{expOnNonNeuralModels}
These models deal with classification, meaning the model must decide between 4 grades instead of selecting a value from [0,3]. TF-IDF word features are used for every model. All models are trained on the train set.

We show results for two types of experiments. When the original sequence and the edit (the new word we are inserting into the title) are provided and when we only provide the edit word. The results can be seen in Table \ref{tab:nonNeuralResults}. These models are not even able to achieve results that can be provided by the simple model that predicts constant (see Table \ref{tab:tableOfSameGradeRes}). Interestingly, comparable results can be achieved with just using the edit word.

\begin{table}[]
\centering
\begin{tabular}{r|c|c|c|c|c|l|c|l|}
\cline{2-9}
\multicolumn{1}{l|}{}               & \multicolumn{4}{c|}{\textbf{Sub-Task 1 [RMSE]}}                                               & \multicolumn{4}{c|}{\textbf{Sub-Task 2 [Accuracy]}}                                                                             \\ \cline{2-9} 
\multicolumn{1}{l|}{}               & \multicolumn{2}{c|}{\textbf{orig. + edit word}} & \multicolumn{2}{c|}{\textbf{edit word}} & \multicolumn{2}{c|}{\multirow{2}{*}{\textbf{orig. + edit word}}} & \multicolumn{2}{c|}{\multirow{2}{*}{\textbf{edit word}}} \\ \cline{2-5}
\multicolumn{1}{l|}{}               & \textbf{all grades}     & \textbf{3. grade}     & \textbf{all grades} & \textbf{3. grade} & \multicolumn{2}{c|}{}                                            & \multicolumn{2}{c|}{}                                    \\ \hline
\multicolumn{1}{|r|}{\textbf{DTC}}  & 0.987                   & 0.917                 & 0.968               & 0.914             & \multicolumn{2}{c|}{0.481}                                       & \multicolumn{2}{c|}{0.520}                               \\ \hline
\multicolumn{1}{|r|}{\textbf{SVM}}  & 0.955                   & \textbf{0.834}        & 0.968               & 0.914             & \multicolumn{2}{c|}{0.514}                                       & \multicolumn{2}{c|}{0.533}                               \\ \hline
\multicolumn{1}{|r|}{\textbf{k-nn}} & 0.998                   & 0.860                 & 1.012               & 0.933             & \multicolumn{2}{c|}{0.542}                                       & \multicolumn{2}{c|}{\textbf{0.543}}                      \\ \hline
\multicolumn{1}{|r|}{\textbf{NBC}}  & 1.574                   & 0.911                 & 1.969               & 1.626             & \multicolumn{2}{c|}{0.482}                                       & \multicolumn{2}{c|}{0.266}                               \\ \hline
\end{tabular}
\caption{\label{tab:nonNeuralResults}Results on the test set for non-neural models. The number of neighbors for k-NN differs among the experiments. For the Sub-Task 1 we use $k=5$, and for Sub-Task 2  $k=13$ is used.}
\end{table}

\subsection{Neural models}
\label{expOnNeuralModels}
In addition to the model used in our submission (and its version JokeMeterBoosted), we performed experiments with a system using a pre-trained ALBERT model (JokeALBERT). JokeALBERT utilizes contextual embeddings of the whole input sequence from ALBERT and then selects those that belong to the edited word and averages them into one. Finally, the linear transformation with softmax is applied.

To all our neural models, we provided input that uses the same format (see section \ref{jokeMeter}). For both models, we use the cross-entropy loss.  We used \textit{Adam} \cite{adamOptim} with weight decay \cite{adamW} as a optimizer. We stop the training after no improvement in RMSE on the dev set in five subsequent epochs. The results for these models are presented in Table \ref{tab:neuralResults}. Results for JokeMeter and JokeALBERT were obtained with batch size 16 and learning rate 1e-5.

\begin{table}[]
\centering
\begin{tabular}{r|c|c|c|c|c|c|c|}
\cline{2-8}
\multicolumn{1}{l|}{}                                    & \textbf{Baseline} & \textbf{Best} & \textbf{JM (all)} & \textbf{JM (3.)} & \textbf{JMB (all)} & \textbf{JA (all)} & \textbf{JA (3.)} \\ \hline
\multicolumn{1}{|r|}{\textbf{Sub-Task 1 {[}RMSE{]}}}     & 0.575             & 0.497         & \textbf{0.558}    & 0.573 & 0.545           & 0.533             & 0.554            \\ \hline
\multicolumn{1}{|r|}{\textbf{Sub-Task 2 {[}Accuracy{]}}} & 0.490             & 0.674         & \textbf{0.578}    & 0.538 & 0.605            & 0.640             & 0.615            \\ \hline
\end{tabular}
\caption{\label{tab:neuralResults}Results on the test set for neural models. We also provide results of the baseline and the best model in the competition. Our official results are in JM (all) column. The JM abbreviation means JokeMeter, JMB is JokeMeterBoosted, JA is JokeALBERT, all means that the \textit{all-grades} training was used and 3. that the \textit{3-grade} training was used.}
\end{table}

\subsection{Evaluation and baseline}
\label{evalBaseline}

All results for trained models were evaluated with the official scripts. For the Sub-Task 1, we always show the root-mean-square error (RMSE)  metric, and for the Sub-Task 2, the accuracy is used.

The baseline system for Sub-Task 1 always predicts the mean funniness grade from the training set (0.936), and for Sub-Task 2, it always predicts the most frequent label in the training set (1).

\begin{figure}[h]
    \centering
    \includegraphics[width=0.7\linewidth]{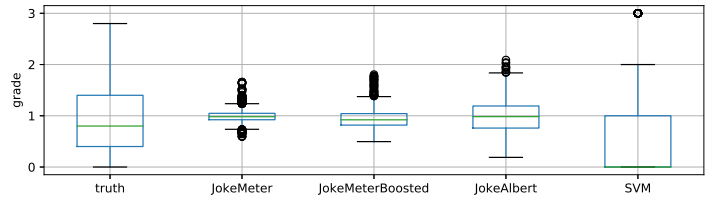}

    \caption{\label{fig:boxPlotOfResults}Comparison of grade predictions for multiple models and the truth labels. We can see that JokeMeter and JokeALBERT's predictions are focused on a small interval around the one. JokeMeter, JokeMeterBoosted and JokeALBERT are for all-grades training, and the SVM is for 3-grade training.}
    
\end{figure}

\section{Conclusion}
\label{conclusion}
The system description was provided, and we compare the achieved results of the official model with several other models, including the baseline and the best team in the competition.

In future work, it should be more investigated if imbalanced dataset and small inter-annotator agreement caused that the JokeMeter model was more focused on the prior probabilities of grades and not on the input itself (see Figure \ref{fig:boxPlotOfResults}). 

\section*{Acknowledgements}

This work was supported by the Czech Ministry of Education, Youth and Sports, subprogram INTERCOST, project code: LTC18054.  

\bibliographystyle{coling}
\bibliography{semeval2020}

\appendix

\section{Supplemental Material}
\label{supplemental}
All the results presented in this section are averages from 3 runs. The evaluation is always done on the dev set.

\subsection{Embeddings}
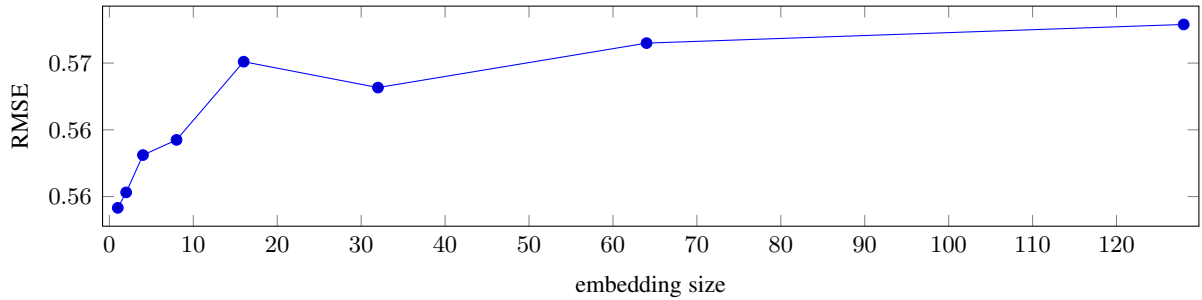
\begin{figure}[H]
    \centering
    \input{images/embeddings_sensitivity.tex}

    \caption{\label{fig:embeddingsSensitivity}Influence of the size of the token embedding on the RMSE.}
    
\end{figure}

In Figure \ref{fig:embeddingsSensitivity} can be seen the influence of the token embedding size on the RMSE. The rest of the JokeMeter model configuration remains the same as described in section \ref{jokeMeter}.

\subsection{Convolutional features}
\begin{figure}[H]
    \centering
    \input{images/convolutional_features_sensitivity.tex}

    \caption{\label{fig:convFeaturesSensitivity}Influence of the number of filters per region size on the RMSE.}
    
\end{figure}
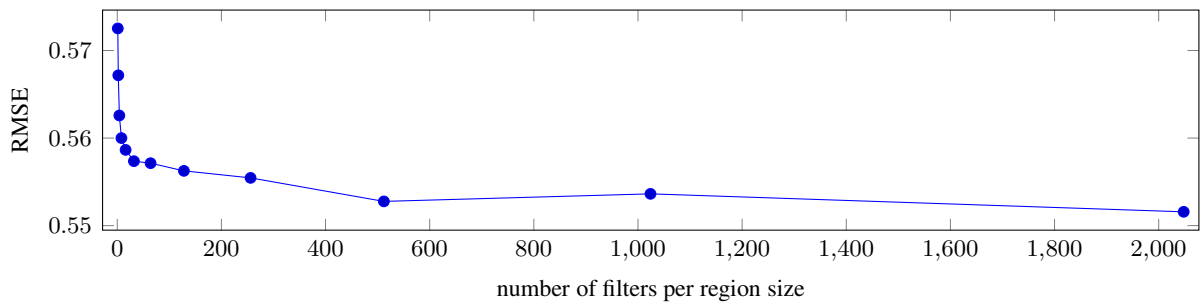

In Figure \ref{fig:convFeaturesSensitivity} can be seen the influence of the used convolutional filters per region size on the RMSE. The rest of the JokeMeter model configuration remains the same as described in section \ref{jokeMeter}.

\begin{figure}[H]
    \centering
    \input{images/conv_2048_embeddings_sensitivity.tex}

    \caption{\label{fig:embeddingsSensitivityFor2048}Influence of the size of the token embedding on the RMSE for 2048 convolutional filters per region size.}
    
\end{figure}
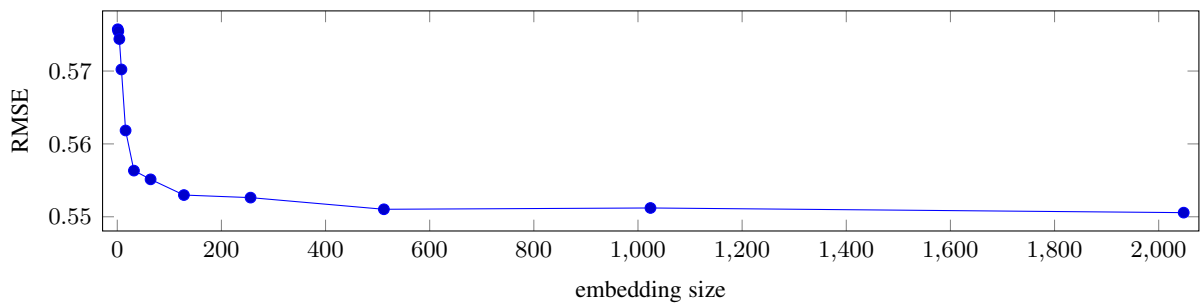

As shown in Figure \ref{fig:embeddingsSensitivityFor2048} there is a different relation between token embedding size and RMSE for 2048 convolutional filters per region size than for the default 2 (see Figure \ref{fig:embeddingsSensitivity}).

\subsection{Ablation experiments}

In this section are presented results for ablation experiments for JokeMeter that uses 2048 convolutional filters per region size and
 2048 size of the token embeddings. The rest of the model configuration remains the same as described in section \ref{jokeMeter}.
 
The influence of used features summarizes Table \ref{tab:ablation}. We can see that the usage of edit embedding does not improve results. We used these findings to create a model that is using 2048 convolutional filters per region size, 2048 dimensional token embeddings, and no edit embedding; we call it JokeMeterBoosted. 

\begin{table}[H]
\centering
\begin{tabular}{r|c|}
\cline{2-2}
\multicolumn{1}{l|}{}                                                    & \textbf{RMSE} \\ \hline
\multicolumn{1}{|r|}{\textbf{convolutional features only}}               &      0.550260959621652 $\pm$ 0.0012         \\ \hline
\multicolumn{1}{|r|}{\textbf{edit embedding only}}                       & 0.63520130408731 $\pm$ 0.0005           \\ \hline
\multicolumn{1}{|r|}{\textbf{convolutional features and edit embedding}} & 0.5505674648279042 $\pm$ 0.0008             \\ \hline
\end{tabular}
\caption{\label{tab:ablation} Results for the ablation experiments on the JokeMeterBoosted.}
\end{table}
 
\subsection{Batch size and learning rate analysis for JokeMeterBoosted}
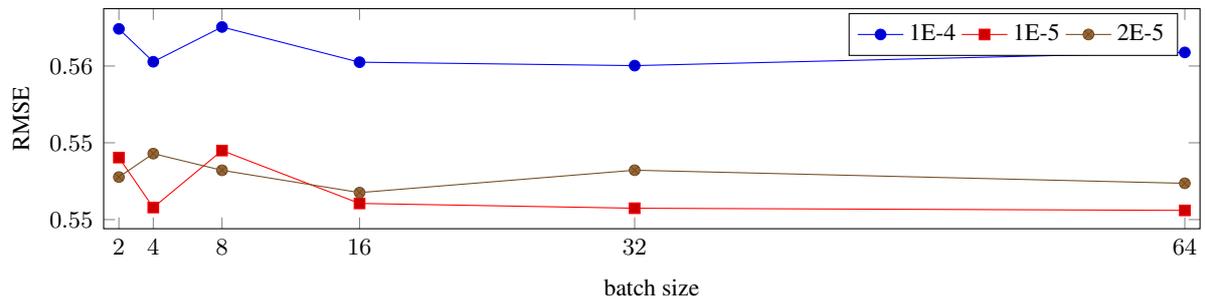
\begin{figure}[H]
    \centering
    \input{images/batch_size_learning_rate.tex}

    \caption{\label{fig:batchSizeLearningRate}JokeMeterBoosted RMSE for three learning rates depending on the batch size.}
    
\end{figure}

According to Figure \ref{fig:batchSizeLearningRate} we used for the JokeMeterBoosted batch size 64 and learning rate 1E-5.

\end{document}

%% file: images/histogram_mean_imbalanced.tex
\begin{tikzpicture}[font=\small]
\pgfplotsset{width=\textwidth, height=4.0cm,}
\begin{axis}[
    xlabel={mean grade},
    ylabel={samples},
    symbolic x coords={0.0-0.3, 0.3-0.6, 0.6-0.9, 0.9-1.2, 1.2-1.5, 1.5-1.8, 1.8-2.1, 2.1-2.4, 2.4-2.7, 2.7-3.0, 3.0},
    xticklabel style={rotate=90},
    bar width=0.2cm,
    xlabel style={yshift=-0.55cm},
    xtick=data]
    
    \addplot[ybar,fill=black, opacity=0.8] coordinates {
       (0.0-0.3, 1451)
       (0.3-0.6, 1106)
       (0.6-0.9, 2357)
       (0.9-1.2, 1173)
       (1.2-1.5, 1850)
       (1.5-1.8, 650)
       (1.8-2.1, 744)
       (2.1-2.4, 173)
       (2.4-2.7, 133)
       (2.7-3.0, 13)
       (3.0, 2)
    };
    
\end{axis}
\end{tikzpicture}

%% file: images/imbalancement_per_position.tex
\begin{tikzpicture}[font=\small]
\pgfplotsset{width=\textwidth, height=4.5cm,}

\begin{axis}[
    xlabel={grade},
    ylabel={samples},
    ylabel near ticks,
    legend style={legend columns=-1, nodes={scale=1.0, transform shape}}, 
    enlarge x limits={abs=0.2cm},
    xtick=data]
    
    \addplot coordinates {
       (0, 20439)
       (1, 14303)
       (2, 9208)
       (3, 4310)
    };
    \addlegendentry{any}
    
    \addplot coordinates {
        (0, 523)
        (1, 2296)
        (2, 3657)
        (3, 3176)
    };
    \addlegendentry{1.}
    
    \addplot coordinates {
        (0, 1896)
        (1, 3816)
        (2, 3071)
        (3, 869)
    };
    \addlegendentry{2.}
    
    \addplot coordinates {
        (0, 3744)
        (1, 4030)
        (2, 1669)
        (3, 209)
    };
    \addlegendentry{3.}
    
    \addplot coordinates {
        (0, 6082)
        (1, 2875)
        (2, 652)
        (3, 43)
    };
    \addlegendentry{4.}
    
    \addplot coordinates {
        (0, 8194)
        (1, 1286)
        (2, 159)
        (3, 13)
    };
    \addlegendentry{5.}
    
\end{axis}
\end{tikzpicture}

%% file: images/embeddings_sensitivity.tex
\begin{tikzpicture}[font=\small]
\pgfplotsset{width=\textwidth, height=4.5cm,}

\begin{axis}[
    xlabel={embedding size},
    ylabel={RMSE},
    ylabel near ticks,
    legend style={legend columns=-1, nodes={scale=1.0, transform shape}}, 
    enlarge x limits={abs=0.2cm}]
    
    \addplot coordinates {
       (1, 0.5616621856746101)
       (2, 0.5621261460275376)
       (4, 0.5632435029135926)
       (8, 0.5636987181226335)
       (16, 0.5660412267076357)
       (32, 0.5652657362569217)
       (64, 0.5665980208467153)
       (128, 0.5671569491448157)
    };
    
\end{axis}
\end{tikzpicture}

%% file: images/convolutional_features_sensitivity.tex
\begin{tikzpicture}[font=\small]
\pgfplotsset{width=\textwidth, height=4.5cm,}

\begin{axis}[
    xlabel={number of filters per region size},
    ylabel={RMSE},
    ylabel near ticks,
    legend style={legend columns=-1, nodes={scale=1.0, transform shape}}, 
    enlarge x limits={abs=0.2cm}]
    
    \addplot coordinates {
       (1, 0.5725216571115604)
       (2, 0.5671630219737295)
       (4, 0.5625775251607559)
       (8, 0.5599970177983083)
       (16, 0.5586478222292511)
       (32, 0.5573606407764417)
       (64, 0.5571281672211628)
       (128, 0.5562522583577061)
       (256, 0.5554518568372412)
       (512, 0.5527660290080281)
       (1024, 0.5536271019414571)
       (2048, 0.5515763131356543)
    };
    
\end{axis}
\end{tikzpicture}

%% file: images/conv_2048_embeddings_sensitivity.tex
\begin{tikzpicture}[font=\small]
\pgfplotsset{width=\textwidth, height=4.5cm,}

\begin{axis}[
    xlabel={embedding size},
    ylabel={RMSE},
    ylabel near ticks,
    legend style={legend columns=-1, nodes={scale=1.0, transform shape}}, 
    enlarge x limits={abs=0.2cm}]
    
    \addplot coordinates {
       (1, 0.5757363671371388)
       (2, 0.5754586966963312)
       (4, 0.5743922378420877)
       (8, 0.570218298698682)
       (16, 0.5618431525214714)
       (32, 0.5563338753483579)
       (64, 0.5551361709432585)
       (128, 0.5529816045374728)
       (256, 0.5526283906301224)
       (512, 0.5510288290948041)
       (1024, 0.551207806461249)
       (2048, 0.5505674648279042)
    };
    
\end{axis}
\end{tikzpicture}

%% file: images/batch_size_learning_rate.tex
\begin{tikzpicture}[font=\small]
\pgfplotsset{width=\textwidth, height=4.5cm,}

\begin{axis}[
    xlabel={batch size},
    ylabel={RMSE},
    ylabel near ticks,
    legend style={legend columns=-1, nodes={scale=1.0, transform shape}}, 
    enlarge x limits={abs=0.2cm},
    xtick=data]
    
    \addplot coordinates {
       (2, 0.5624171626104076)
       (4, 0.5602771271640967)
       (8, 0.5625403724970804)
       (16, 0.5602480234508137)
       (32, 0.5600255345961422)
       (64, 0.5608856621095183)
    };
    \addlegendentry{1E-4}
    
    \addplot coordinates {
       (2, 0.5540370508050744)
       (4, 0.5507843882910859)
       (8, 0.5544943088226374)
       (16, 0.5510516151082169)
       (32, 0.5507397098080787)
       (64, 0.5505990141685929)
    };
    \addlegendentry{1E-5}
    
    \addplot coordinates {
       (2, 0.5527685045963303)
       (4, 0.5542955082921143)
       (8, 0.5532087339037847)
       (16, 0.5517581548482565)
       (32, 0.5532148174845312)
       (64, 0.5523622821831925)
    };
    \addlegendentry{2E-5}
    
\end{axis}
\end{tikzpicture}